%% file: main.tex
\documentclass[conference]{IEEEtran}
\input{packages}
\input{commands}

\IEEEoverridecommandlockouts
\usepackage{graphicx}
\def\BibTeX{{\rm B\kern-.05em{\sc i\kern-.025em b}\kern-.08em
    T\kern-.1667em\lower.7ex\hbox{E}\kern-.125emX}}
\usepackage[font=small,labelfont=bf,tableposition=top]{caption}

\usepackage{blindtext}
\title{here title}

\let\oldtwocolumn\twocolumn
\renewcommand\twocolumn[1][]{%
    \oldtwocolumn[{#1}{
    \vspace{-0.5cm}
    \begin{center}
\includegraphics[width=.7\linewidth]{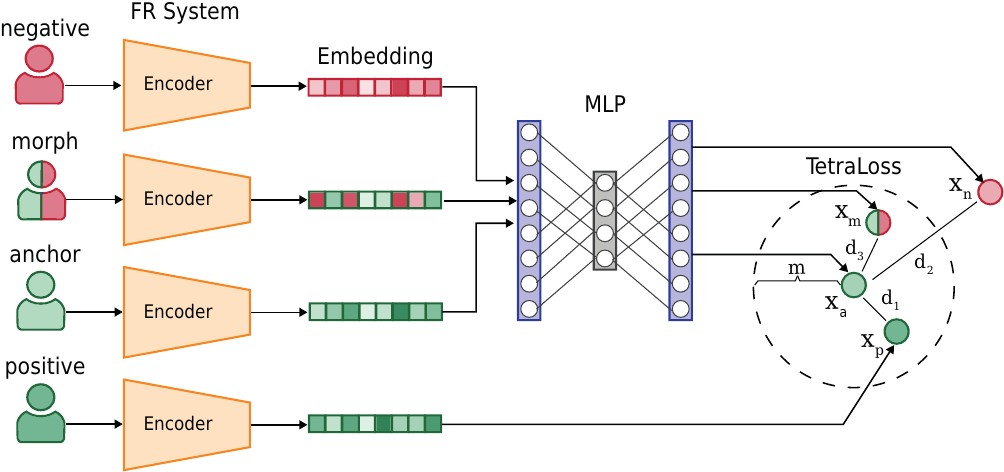}
\captionof{figure}{Overview of the proposed TetraLoss system for finetuning embeddings of a face recognition system to be more robust against morphing attacks. Embeddings corresponding to anchor, positive, negative and morph classes are extracted using a pre-trained face recognition system whereafter the embeddings are fed to a multilayer perceptron (MLP) to obtain new embeddings. The MLP is trained using the proposed TetraLoss function where it learns to achieve low intra-class variation by lowering the distance ($d_1$) between the anchor and positive embeddings ($x_a$, $x_p$) while pushing the morph ($x_m$) and negative ($x_n$) embeddings away from the anchor beyond some margin ($m$).}
\label{fig:arch_overview}
        \end{center}
    }]
}

\title{\LARGE \bf
TetraLoss: Improving the Robustness of Face Recognition against Morphing Attacks
}

\author{\parbox{16cm}{\centering
    {\large Mathias Ibsen$^1$ and L. J. Gonzalez-Soler$^1$ and Christian Rathgeb$^1$ and Christoph Busch$^1$}\\
    {\normalsize
    $^1$ da/sec---Biometrics and Security Research Group, Hochschule Darmstadt, 64295 Darmstadt, Germany}} %
}

\newcommand\copyrighttext{%
  \footnotesize \textcopyright 2024 IEEE. Personal use of this material is permitted. Permission from IEEE must be obtained for all other uses, in any current or future media, including reprinting/republishing this material for advertising or promotional purposes, creating new collective works, for resale or redistribution to servers or lists, or reuse of any copyrighted component of this work in other works.
  DOI: \href{https://doi.org/10.1109/FG59268.2024.10581988}{10.1109/FG59268.2024.10581988}}
\newcommand\copyrightnotice{%
\begin{tikzpicture}[remember picture,overlay]
\node[anchor=south,yshift=10pt] at (current page.south) {\fbox{\parbox{\dimexpr\textwidth-\fboxsep-\fboxrule\relax}{\copyrighttext}}};
\end{tikzpicture}%
}

\begin{document}

\maketitle
\copyrightnotice
\IEEEpubidadjcol

\begin{abstract}
Face recognition systems are widely deployed in high-security applications such as for biometric verification at border controls. Despite their high accuracy on pristine data, it is well-known that digital manipulations, such as face morphing, pose a security threat to face recognition systems. Malicious actors can exploit the facilities offered by the identity document issuance process to obtain identity documents containing morphed images. Thus, subjects who contributed to the creation of the morphed image can with high probability use the identity document to bypass automated face recognition systems. In recent years, no-reference (\ie single image) and differential morphing attack detectors have been proposed to tackle this risk. These systems are typically evaluated in isolation from the face recognition system that they have to operate jointly with and do not consider the face recognition process. Contrary to most existing works, we present a novel method for adapting deep learning-based face recognition systems to be more robust against face morphing attacks. To this end, we introduce TetraLoss, a novel loss function that learns to separate morphed face images from its contributing subjects in the embedding space while still achieving high biometric verification performance. In a comprehensive evaluation, we show that the proposed method can significantly enhance the original system while also significantly outperforming other tested baseline methods.
\end{abstract}

\section{INTRODUCTION}
Face recognition systems have become ubiquitous in private and commercial applications such as unlocking smartphones or automated access control systems. The accuracy and usefulness of these systems have increased rapidly in recent years with advancements in deep learning-based techniques and accessibility to large training datasets. Nowadays, such automated systems achieve near-perfect biometric performance on many challenging benchmarks~\cite{Grother-FRTechnologyEvaluationPart1Verification-FRTE-2023,Kim-AdaFace-CVPR-2022,Deng-ArcFace-IEEE-CVPR-2019}, \eg on the LFW~\cite{Huang-LFW-WorkshopRealLifeImages-2008} and CFP-FP~\cite{Sengupta-CFPFP-WACV-2016} datasets. In order to achieve high biometric performance, such systems must learn to generalize to unseen data, be robust to intra-subject variations, and invariant towards various environmental factors such as differences in illumination and varying backgrounds. 

\begin{figure}[!tb]%
\centering
\begin{subfigure}[t]{0.3\linewidth}
  \includegraphics[width=\textwidth]{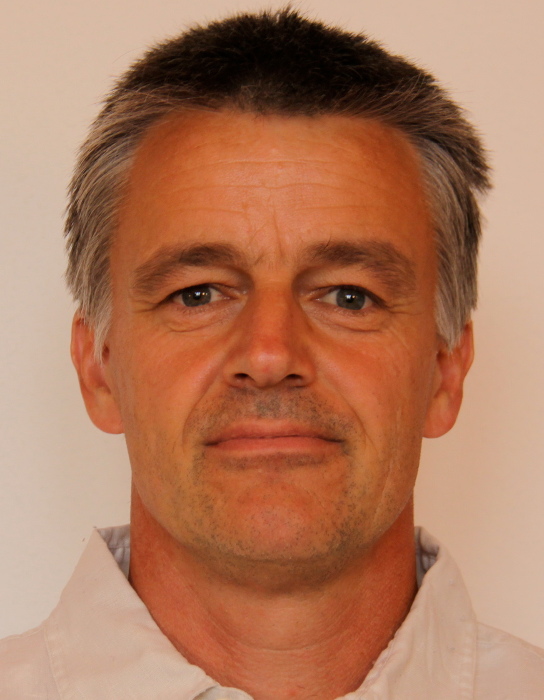}
        \caption{Subject 1}
\end{subfigure}\quad %
\begin{subfigure}[t]{0.3\linewidth}
  \includegraphics[width=\textwidth]{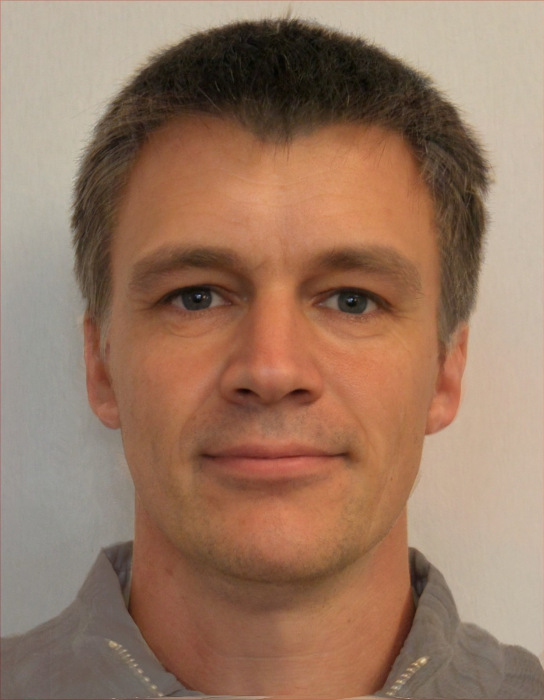}
        \caption{Morph}
\end{subfigure}\quad %
\begin{subfigure}[t]{0.3\linewidth}
  \includegraphics[width=\textwidth]{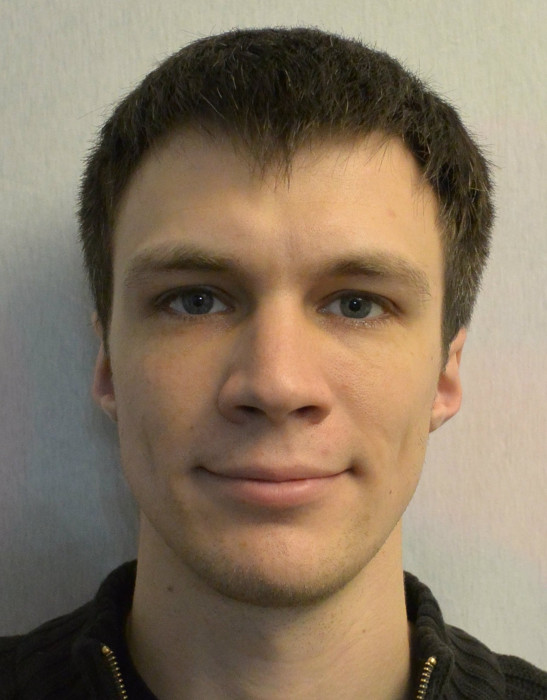}
  \caption{Subject 2}
\end{subfigure}
    \caption{Examples of a morphed face image (b) created from subject 1 (a) and subject 2 (c). Taken from~\cite{Scherhag-FaceMorphingAttacks-TIFS-2020}.}
\label{fig:good_face_morph}
\end{figure}

Despite the advantages of facial recognition systems, their increased performance and generalisation have led to an intensification of their vulnerability to a variety of physical presentation attacks~\cite{Raghavendra-FacePAD-Survey-2017}, \eg silicone masks, as well as digital attacks~\cite{Ibsen-DigitalFaceManipulationsBiometricSystems-Handbook-2022}. Especially face morphing attacks, \ie a digital manipulation where a malicious actor combines the face of two individuals into a single image (see~\autoref{fig:good_face_morph}), have been shown to pose a realistic security challenge towards automated border control systems (also called eGates)~\cite{Scherhag-FaceRecognitionVulnarabilityTowardsMorphedFace-IWBF-2017}. In such systems, biometric information stored in an electronic Machine Readable Travel Document (eMRTD) is used together with a live capture to perform face verification automatically. In many countries, the issuing process allows individuals to submit their own face images when applying for such an eMRTD, which are then checked manually by the issuing authority. Since morphed face images cannot reliably be detected by humans~\cite{Nichols-HumanPerformanceDetectingDigitalFaceImageManipulations-ACCESS-2022, Rathgeb-CrowdManipulationDetection-ICIP-2022}, it has been shown that it is possible to get issued an eMRTD containing a morphed image. In such cases and for high-quality morphs, the eMRTD can, with high probability, be used by multiple individuals to bypass the automated control systems. Currently, we are observing the morphing attack paradox, \ie the better the face recognition system becomes in terms of biometric performance and the lower its error rates on bona fide mated comparison trials, the more vulnerable the face recognition system becomes to morphing attacks, \eg as shown in~\cite{Ngan-UtilityOfFRAlgorithmsForMorphDetection-FRVT-2022}.

To mitigate the security risk posed by morphing attacks, the European Regulation 2019/1157~\cite{EU-Regulation-2019-1157-on-CitizenCard-and-LiveEnrolment-190620} recommended in recital 32 that the issuing authorities of identity cards in member states should consider collecting facial images during a live enrolment process. Unfortunately, this policy is currently only adopted by a few EU member states; hence, it is still possible to submit such morphed images in many countries, and old identity documents are still in circulation. Therefore, countermeasures need to be built into face recognition systems to make them more robust against morphing attacks. To this end, several morphing attack detection (MAD) algorithms for detecting morphed face images have been proposed (see, \eg~\cite{Ibsen-DifferentialAnomalyDetectionForFacialImages-WIFS-2021,Scherhag-MorphingSurvey-IEEE-2019}). Of these, particularly the so-called differential morphing attack detection (D-MAD) methods have shown prominence. These approaches detect morphing attacks by considering both a trusted image (\eg live capture) and a suspected image (\eg passport image), which are available during authentication in biometric verification systems. Despite significant improvements of such detection systems, recent public benchmarks show that they cannot achieve sufficient detection performance at relevant operational high-security thresholds~\cite{NGAN-PerformanceOfAutomatedFaceMorphDetection-FRVT-2020}.
Furthermore, most benchmarks neglect to evaluate such detection systems in a full-system evaluation where the detection system operates in conjunction with a biometric verification system. In such a realistic scenario the detection and verification system must jointly assess the authenticity of the passport image and verify the identity claim. An alternative to such systems is to make the face recognition system aware of morphing attacks by considering morphed images during the training of the face recognition model. Such a system offers a more seamless evaluation as it can be evaluated considering only the biometric verification performance. 

In this work, we propose a novel mechanism to modify the embedding space of an existing deep learning-based face recognition system and make them more robust against morphing attacks. This has the advantage that both the original face recognition embeddings and the modified embeddings can be considered, while still maintaining good biometric performance. Furthermore, such a modified face recognition system can still be combined with traditional stand-alone detection algorithms. The experimental evaluation of our proposed approach is only focused on the use case of morphing attacks. However, the method can likely be extended for other digital or physical manipulations. Figure~\ref{fig:arch_overview} shows an overview of the proposed method which is referred to as \emph{TetraLoss}\footnote{\url{https://github.com/dasec/TetraLoss}}.

In summary, this work makes the following contributions:

\begin{itemize}
    \item A framework that can be placed on top of existing deep learning-based face recognition systems to make them more robust against morphing attacks.
    \item A new loss function called \emph{TetraLoss} that can be used as an optimisation function for face recognition systems during training to separate morph embeddings from the embeddings of their contributing subjects.
    \item A systematic and comprehensive evaluation of the proposed method compliant with the metrics defined in the international standard ISO/IEC 30107-3~\cite{ISO-IEC-30107-3-PAD-metrics-2023}.
\end{itemize}

The remainder of the paper is structured as follows: Section~\ref{sec:related_work} introduces relevant related work regarding vulnerability studies of face recognition systems towards morphing attacks and face MAD. Section~\ref{sec:proposed_framework} describes the framework for adapting existing face recognition systems to be more robust against morphing attacks. Section~\ref{sec:experimental_setup} and~\ref{sec:generalized_system_evaluation} describe the experimental setup and results for a generalized full-system evaluation, respectively. Section~\ref{sec:feasibility_results} investigates the feasibility of the proposed loss function when compared to triplet loss. Lastly, Section~\ref{sec:conclusion} summarizes the proposed method and findings of the paper. 

\section{RELATED WORK}
\label{sec:related_work}
\subsection{Face Recognition Systems under Morphing Attacks}
The vulnerability of face recognition systems to morphing attacks was first demonstrated in~\cite{Ferrara-TheMagicPassport-IJCB-2014}, where the authors showed that a criminal, with the help of an accomplice, can bypass the security of an eGate using a morphed face image. In practice, a face morph is created by combining the facial representations of two faces into a single face image. The accomplice then applies for an eMRTD using the morphed face image. Once the document is issued, the accomplice can use the image to bypass the eGate. Since this first work, several other works have shown the vulnerability of state-of-the-art face recognition systems to morphing attacks, \eg~\cite{Scherhag-FaceRecognitionVulnarabilityTowardsMorphedFace-IWBF-2017, Scherhag-MorphingAttacks-MorphingTechniques-BIOSIG-2017, Sarkar-AreGANsThreateningFR-ICASSP-2022, Ferrara-MorphingAttackPotential-IWBF-2022}. Early works, \eg~\cite{Scherhag-FaceRecognitionVulnarabilityTowardsMorphedFace-IWBF-2017, Scherhag-MorphingAttacks-MorphingTechniques-BIOSIG-2017} focused mainly on landmark-based facial morphs where facial landmarks are detected on each source image and used for blending and warping to generate the morphed facial images. Other studies have also looked at GAN-based~\cite{Sarkar-AreGANsThreateningFR-ICASSP-2022} and diffusion-based~\cite{Damer-MorDIFF-IWBF-2023} morphs. An overview of works investigating the vulnerability of morphs on different face recognition systems is shown in~\autoref{tab:fr_vulnerability_to_morphs}. The table primarily reports works where vulnerability to new morphing attacks or databases is shown. Hence, works or benchmarks that focus on MAD are not reported; here, the reader is instead referred to~\cite{Raja-SOTAMD-TIFS-2021,Venkatesh-FaceMorphingAttackGenerationAndDetection-TTS-2021} for more comprehensive overviews of morphing databases. 

To evaluate the vulnerability to morphing attacks, different metrics have been proposed. In~\cite{Scherhag-MorphingAttacks-MorphingTechniques-BIOSIG-2017}, the authors proposed the Mated Morph Presentation Match Rate (MMPMR), along with two variants in case of multiple images per subject, where a morphing attack is considered successful if all contributing subjects match against the morphed sample. In~\cite{Venkatesh-OnInfluenceOfAgeingOnFaceMorphAttacks-IJCB-2020}, the authors proposed a more restrictive measure which requires successful verification for all contributing subjects and all attempts. In~\cite{Ferrara-MorphingAttackPotential-IWBF-2022}, the authors proposed to extend these existing metrics by considering multiple face recognition systems under the motivation that a good morphing attack should potentially be able to fool multiple state-of-the-art face recognition systems. A problem with these metrics is that they only evaluate the vulnerability of the face recognition system to morphs independent of the false reject rate of the system. Hence, in~\cite{Scherhag-MorphingAttacks-MorphingTechniques-BIOSIG-2017}, the authors proposed the Relative Impostor Attack Presentation Accept Rate (RIAPAR)~\footnote{Previously called Relative Morph Match Rate (RMMR)} which at a given security threshold considers both the attack success rate and the false non-match rate of the system. RIAPAR was recently introduced into the international standard ISO/IEC 30107-3~\cite{ISO-IEC-30107-3-PAD-metrics-2023} and hence, in compliance with this, will be reported in this paper. 

In~\cite{Kelly-ImprovingRobustnessofFRtoMorphingAttacks-SPPR-2020}, the authors presented the first work which considered morphing attacks during the development of a face recognition system. Specifically, the authors use an existing convolutional neural network with pre-trained weights and retrain the last fully connected layer using triplet loss. Since the morphed images can contain visible artefacts, the authors propose to add similar artefacts to mated samples which is achieved by augmenting the mated pairs by utilizing the same morphing technique on image-pairs stemming from the same subject. During training, different triplet combinations of the anchor, negative and positive classes are selected in order to achieve increased robustness to morphing attacks. 

\begin{table}[!t]
    \centering
    \caption{Overview of papers and used morphed databases from the literature which shows vulnerability of face recognition systems to morphing attacks. LM indicates landmark-based morphs and $*$ indicates that vulnerability was shown using at least one commercial face recognition system.}
\begin{adjustbox}{max width=\linewidth}
    \begin{tabular}{@{}lllllp{3.1cm}@{}} \toprule  \textbf{Paper}  & \textbf{Morphing method} & \textbf{Morph Types} & \textbf{Print-scan} & \textbf{\#Morphs}  & \textbf{Face recognition system} \\ \midrule
    \multirow{2}{*}{\cite{Ferrara-TheMagicPassport-IJCB-2014}\textsuperscript{\raisebox{0.3ex}{$*$}}} & \multirow{2}{*}{GIMP GAP} & \multirow{2}{*}{LM} & \multirow{2}{*}{No} & \multirow{2}{*}{12} &  VeriLookSDK \& Luxand FaceSDK  \\   \midrule
    \multirow{2}{*}{\cite{Ferrara-FaceImageAlterations-Springer-2016}\textsuperscript{\raisebox{0.3ex}{$*$}}}  & \multirow{2}{*}{GIMP GAP} & \multirow{2}{*}{LM} & \multirow{2}{*}{No} & \multirow{2}{*}{21} &  VeriLookSDK \& Luxand FaceSDK \\\midrule
     \multirow{1}{*}{\cite{Raghavendra-DetectingMorphedFace-BTAS-2016}\textsuperscript{\raisebox{0.3ex}{$*$}}}  & \multirow{1}{*}{GIMP GAP} & \multirow{1}{*}{LM} & \multirow{1}{*}{No} & \multirow{1}{*}{450} &  VeriLookSDK\\ \midrule
     \multirow{2}{*}{\cite{Scherhag-FaceRecognitionVulnarabilityTowardsMorphedFace-IWBF-2017}\textsuperscript{\raisebox{0.3ex}{$*$}}}  & \multirow{2}{*}{GIMP GAP} & \multirow{2}{*}{LM} & \multirow{2}{*}{Yes} & \multirow{2}{*}{231} &  VeriLookSDK \& OpenFace  \\   \midrule
    \multirow{1}{*}{\cite{Raghavendra-FaceMorphingVersusFaceAveraging-IJCB-2017}\textsuperscript{\raisebox{0.3ex}{$*$}}}  & \multirow{1}{*}{-} & \multirow{1}{*}{LM} & \multirow{1}{*}{Yes} & \multirow{1}{*}{1423} &  Cognitec\\ \midrule
     \multirow{1}{*}{\cite{Damer-MorGAN-BTAS-2018}}  & \multirow{1}{*}{MorGAN \& OpenCV} & \multirow{1}{*}{LM \& GAN} & \multirow{1}{*}{No} & \multirow{1}{*}{2000} &  VGG-Face \& OpenFace \\ \midrule
        \multirow{2}{*}{\cite{Scherhag-PRNU-TBIOM-2019}\textsuperscript{\raisebox{0.3ex}{$*$}}}  & OpenCV, FaceMorpher & \multirow{2}{*}{LM} & \multirow{2}{*}{Yes} & \multirow{2}{*}{5972} &  \multirow{2}{*}{COTS} \\ 
         & FaceFusion, UBO Morpher &  & &  &  \\  \midrule
          \multirow{1}{*}{\cite{Ferrara-TextureBlendingAndShapeWarpingInFaceMorphing-IEEE-BIOSIG-2019}\textsuperscript{\raisebox{0.3ex}{$*$}}}  & \multirow{1}{*}{dlib + triangulation} & \multirow{1}{*}{LM} & \multirow{1}{*}{No} & \multirow{1}{*}{560} &  Two COTS systems \\ \midrule
          
          \multirow{1}{*}{\cite{Venkatesh-CanGANMorphsThreatenFR-IWBF-2020}\textsuperscript{\raisebox{0.3ex}{$*$}}}  & \multirow{1}{*}{StyleGAN} & \multirow{1}{*}{GAN} & \multirow{1}{*}{No} & \multirow{1}{*}{2500} &  COTS + ArcFace \\ \midrule
          
        \multirow{2}{*}{\cite{Sarkar-AreGANsThreateningFR-ICASSP-2022}\textsuperscript{\raisebox{0.3ex}{$*$}}}  & OpenCV, FaceMorpher & \multirow{2}{*}{LM \& GAN} & \multirow{2}{*}{No} & \multirow{2}{*}{4888} &  \multirow{2}{*}{COTS} \\ 
         & StyleGAN2, WebMorpher &  & &  &  \\ \midrule 
    \multirow{1}{*}{\cite{Venkatesh-OnInfluenceOfAgeingOnFaceMorphAttacks-IJCB-2020}\textsuperscript{\raisebox{0.3ex}{$*$}}}  & \multirow{1}{*}{UBO Morpher} & \multirow{1}{*}{LM} & \multirow{1}{*}{No} & \multirow{1}{*}{14305} &  Two COTS systems \\ \midrule 

            \multirow{2}{*}{\cite{Damer-MorDIFF-IWBF-2023}}  & \multirow{2}{*}{MorDiff}  & \multirow{2}{*}{Diffusion} & \multirow{2}{*}{No} & \multirow{2}{*}{1000} &  \multirow{1}{*}{ElasticFace, CurricularFace} \\ 
         & &  & &  & MixFaceNet \& PocketNet   \\ 
    \bottomrule
    \end{tabular}
    \label{tab:fr_vulnerability_to_morphs}
    \end{adjustbox}  \vspace{-0.4cm}
\end{table}

\subsection{Detection of Face Morphing Attacks}
Several detection methods have been proposed to increase the security of face recognition systems against morphing attacks. These can be categorized as either being Single Morphing Attack Detection (S-MAD)\footnote{Also called no-reference detection methods} or Differential Morphing Attack Detection (D-MAD). The latter generally achieve the most competitive performance~\cite{NGAN-PerformanceOfAutomatedFaceMorphDetection-FRVT-2020,Scherhag-MorphingSurvey-IEEE-2019}. Several works have proposed such D-MAD algorithms using either handcrafted features, \eg~\cite{Chaudhary-DifferentialMorphFaceDetectionUsingDiscriminativeWavelesSubbands,Scherhag-LandmarkMAD-ICISP-2018,Scherhag-MorphingDetection-DAS-2018} or deep learning-based approaches~\eg~\cite{Borgi-DoubleSiameseMAD-Sensors-2021, Scherhag-FaceMorphingAttacks-TIFS-2020}. An alternative approach to detecting morphed images is face demorphing, where a live face image acquired at an eGate is used to revert part of the morphing process and reveal the legitimate document owner. This idea was first explored in~\cite{Ferrara-Demorphing-TIFS-2018} and has since been used in other work, \eg in~\cite{Ferrara-DemorphingRobustToFacialVariations-EUSIPCO-2018} where the authors improved the robustness of demorphing to face alterations.

To evaluate the different MAD algorithms, several benchmarks have been proposed, \eg the SOTAMD benchmark~\cite{Raja-SOTAMD-TIFS-2021} and the National Institute of Standards and Technology (NIST) face morph detection test~\cite{NGAN-PerformanceOfAutomatedFaceMorphDetection-FRVT-2020}. While these benchmarks show the high potential of methods for detecting face morphing attacks, they also show the current limitations. For instance, the NIST evaluation shows that all present submitted MAD algorithms fail to operate at acceptable morph detection rates for operationally relevant false match rates.

\section{PROPOSED FRAMEWORK}
\label{sec:proposed_framework}

\subsection{Model Architecture}
Inspired by the work in~\cite{Fadi-SelfRestrainedTLForMaskedFR-PatternRecognition-2022}, we propose to add a shallow neural network which operates on top of embeddings from an existing face recognition system. The model consists of four fully connected (FCN) layers, batch normalization (BN) and the Leaky Rectified Linear Unit (Leaky ReLU) activation function. The input layer and the two hidden layers are followed by BN and the Leaky ReLU activation function whereas the output layer is only followed by BN. Given an embedding of size $n$, the model performs a forward pass of the embedding and produces a new embedding of the same size (\ie $n$ = 512 in this work). This multilayer perception (MLP) can be easily adopted to different embedding shapes introduced by different face recognition systems. It is likely that other architectures are capable of improving upon the results obtained in this work, but this is deliberately not explored in depth, as the intention is to propose a general loss and model architecture. Applying a MLP on top of embeddings extracted from an existing face recognition system can be thought of as a type of adapter where the embeddings are adapted for a new task.

\subsection{TetraLoss}
Altering the embeddings of existing deep learning-based face recognition systems offers a seamless integration into most known state-of-the-art face recognition systems. One way to supervise a new network for modifying these embeddings is to use normal triplet loss. Earlier works have shown that triplet loss can be used for learning discriminative face embeddings~\cite{Schroff-FaceNet-CVPR-2015} and even for adopting embeddings to similar tasks such as for recognition of masked face images~\cite{Fadi-SelfRestrainedTLForMaskedFR-PatternRecognition-2022}. 

Consider a batch $x \in X$ of samples, \ie embeddings, obtained by some existing deep learning-based face recognition model. Furthermore, let $\{x^a_i,x^p_i,x^n_i\} \in X$ be a triplet of samples such that $x^a_i$, $x^p_i$ is the anchor and positive samples, respectively and belonging to the same subject, and $x^n_i$ belongs to a different subject. In this case, for a mini-batch of size N, triplet loss can be expressed as: 

\begin{footnotesize}
\begin{equation}
\resizebox{.925\columnwidth}{!}{$L_{triplet} = \frac{1}{N} \sum_{i=1}^{N} \max\{d(x^a_i, x^p_i) + m - d(x^a_i,x^n_i), 0 \}$}
\label{eq:triplet_loss}
\end{equation}
\end{footnotesize}

\noindent where $d$ is the euclidean distance on normalized embeddings and $m$ is the margin which is set to $3$ in this work.

Applying this loss naively on top of an existing embedding would not improve the performance of the system against morphing attacks. Instead, we can replace the negative sample $x^n_i$ in triplet loss by $x^m_i \in X$ where $x^m_i$ is from a morphed image where one of the contributing subjects is the same subject as for the anchor and positive samples and such that all contributing subjects of the morph are different from the subject of the negative sample. Using the morph embeddings in triplet loss would lead to altering the morph embeddings so that they would look less like the embeddings from the contributing subjects. On the contrary, this methodology could considerably affect the biometric verification performance on bona fide samples, as the network would overfit the morph images by not considering the embeddings of completely different identities during training. To tackle this issue, we propose a new loss function called TetraLoss which considers both the anchor, positive, negative, and morph samples during training. Following similar notation as in~\autoref{eq:triplet_loss}, TetraLoss can be defined as follows: 

\begin{footnotesize}
\begin{equation}
\begin{split}
L_{\text{Tetra}} &= \frac{1}{N} \sum_{i=1}^{N} \max\{d(x^a_i, x^p_i) + m - \\
&\qquad \min\{d(x^a_i,x^n_i), d(x^a_i,x^m_i)\}, 0 \}
\end{split}
\label{eq:tetraloss}
\end{equation}
\end{footnotesize}

As seen above, TetraLoss extends the normal triplet loss by considering the smallest distance between the anchor and the morph or negative embeddings. Thus, bona fide embeddings are pushed away from both negative and morph samples. As shown in~\autoref{fig:arch_overview}, embeddings from the anchor, negative, and morph classes are forwarded through the MLP during training. However, the positive class is not, as the idea is to select these samples so they correspond to images captured live at an eGate, and hence, we know they are not manipulated. 

\section{EXPERIMENTAL SETUP}
\label{sec:experimental_setup}
The experimental evaluation is intended to serve two goals: i) investigate whether the proposed method can increase the robustness of state-of-the-art face recognition systems against morphing attacks by lowering its attack potential (\autoref{sec:generalized_system_evaluation}), and ii) evaluate exhaustively the feasibility of the proposed method by comparing TetraLoss to normal triplet loss (\autoref{sec:feasibility_results}). 

\subsection{Face Recognition Systems}
To show that the proposed method can be applied on top of state-of-the-art face recognition systems, we consider three systems namely ArcFace~\cite{Deng-ArcFace-IEEE-CVPR-2019}, MagFace~\cite{Meng-MagFace-CVPR-2021}, and AdaFace~\cite{Kim-AdaFace-CVPR-2022} based on the ResNet100 and ResNet50 backbones and trained on the MS1Mv2 database. What distinguish these methods are mainly the loss function used to train the backbone models: 

\begin{LaTeXdescription}
  \item[ArcFace] or Additive Angular Margin Loss was proposed in~\cite{Deng-ArcFace-IEEE-CVPR-2019} to solve issues with traditional softmax loss which does not explicitly optimize embeddings to achieve high intra-class and low inter-class similarity. Particularly, the authors propose to use the geodesic distance on the hypersphere to maximize the decision boundary between classes. 
\end{LaTeXdescription}

\begin{LaTeXdescription}
  \item[MagFace] was proposed in~\cite{Meng-MagFace-CVPR-2021} to tackle problems with previous angular margin-based losses which are quality-agnostic and hence can lead to unstable within-class distribution as high-quality face images can stay at the decision boundary whereas lower quality images can be at the class center. This is especially a problem in unconstrained in-the-wild recognition. To solve this, MagFace encodes quality information into the feature representation by considering the feature magnitude as a quality indicator.
\end{LaTeXdescription}

\begin{LaTeXdescription}
  \item[AdaFace] was introduced in~\cite{Kim-AdaFace-CVPR-2022} and similar to MagFace the idea is that the relative importance of easy or hard samples should be adjusted based on the quality of the image. This is achieved by the use of a image quality indicator which use feature norm as a proxy for image quality. 
\end{LaTeXdescription}

\subsection{Databases}

\begin{figure}[!t]
    \begin{subfigure}[b]{0.24\linewidth}
        \centering
        \includegraphics[width=\linewidth]{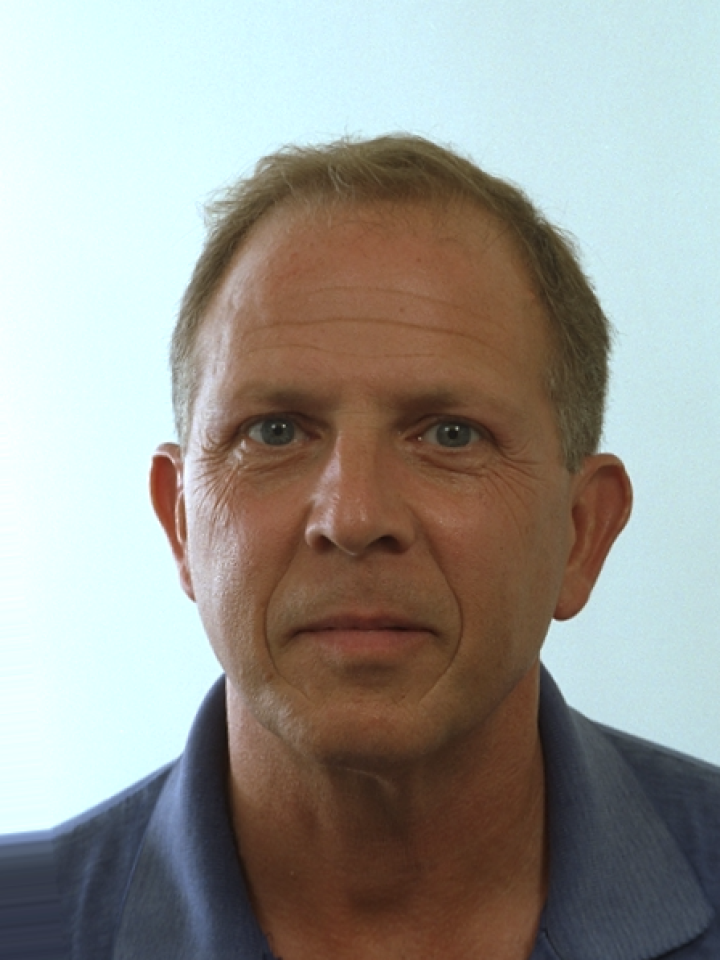}
        \caption[]%
        {{\small reference}}    
        \label{fig:feret_ref_img}
    \end{subfigure}
    \begin{subfigure}[b]{0.24\linewidth}  
        \centering 
        \includegraphics[width=\linewidth]{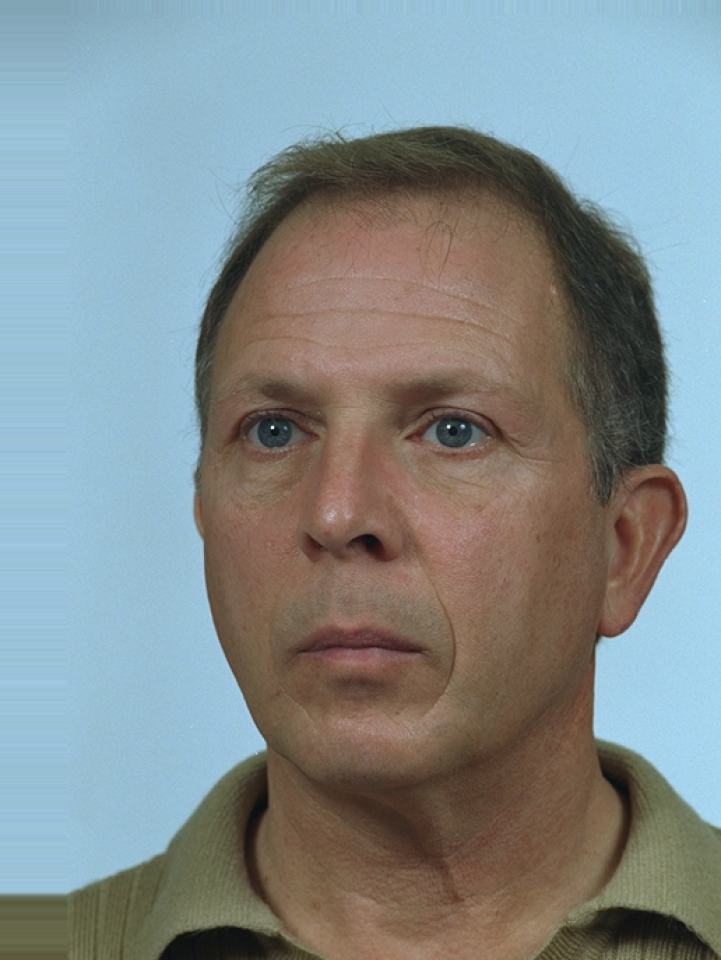}
        \caption[]%
        {{\small probe}} 
         \label{fig:feret_probe_img}
    \end{subfigure}
    \begin{subfigure}[b]{0.24\linewidth}   
        \centering 
        \includegraphics[width=\linewidth]{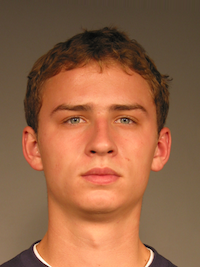}
        \caption[]%
        {{\small reference}}    
    \end{subfigure}
    \begin{subfigure}[b]{0.24\linewidth}   
        \centering 
        \includegraphics[width=\linewidth]{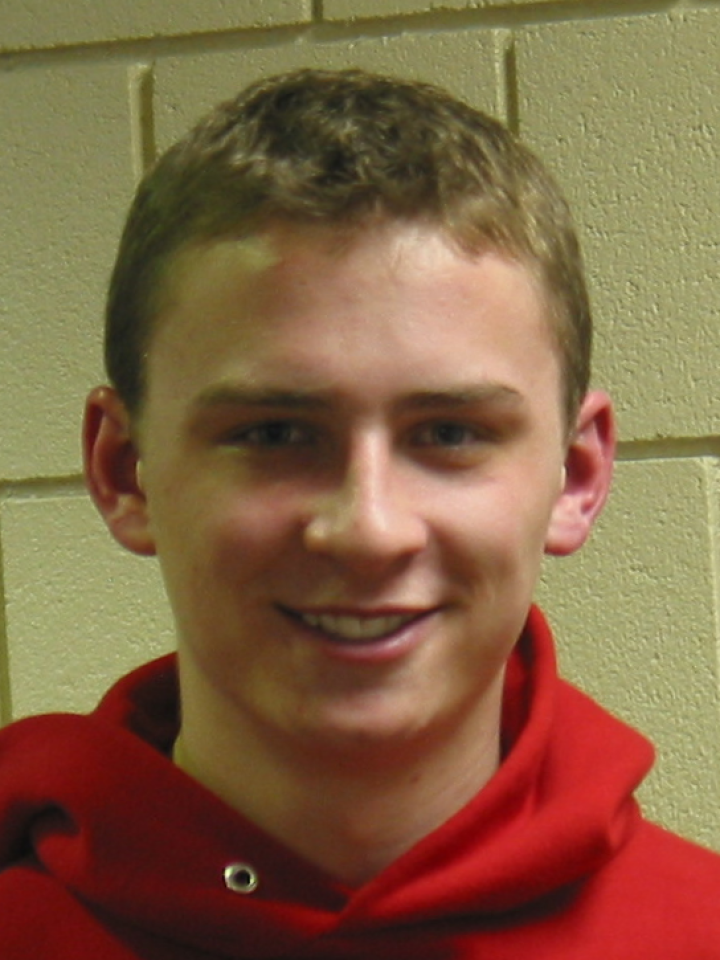}
        \caption[]%
        {{\small probe}}    
    \end{subfigure}
    \caption{Example of reference and probe images from FERET (a, b) and FRGCv2 (c, d).} \vspace{-0.4cm}
    \label{fig:reference_probe_samples}
\end{figure}

For testing the proposed method, the FERET~\cite{Phillips-FERET-1998} and FRGCv2~\cite{Phillips-FRGC-2005} datasets are used. From these datasets morphed images are created using four different morphing tools namely FaceMorpher, OpenCV, FaceFusion, and UBO morpher~\cite{Scherhag-PRNU-TBIOM-2019}. The FERET and FRGCv2 datasets are further split into two subsets such that the reference images are high quality face images and the probe images are more unconstrained face images. This corresponds to the realistic authentication scenario where identification documents often contain more constrained (\ie canonical) face images whereas probe images are captured live by a camera at the scene and hence can be more unconstrained. Examples of reference and probe images for FERET and FRGCv2 are given in~\autoref{fig:reference_probe_samples}.

\begin{figure*}[!tb]
    \centering
    \includegraphics[width=.75\linewidth]{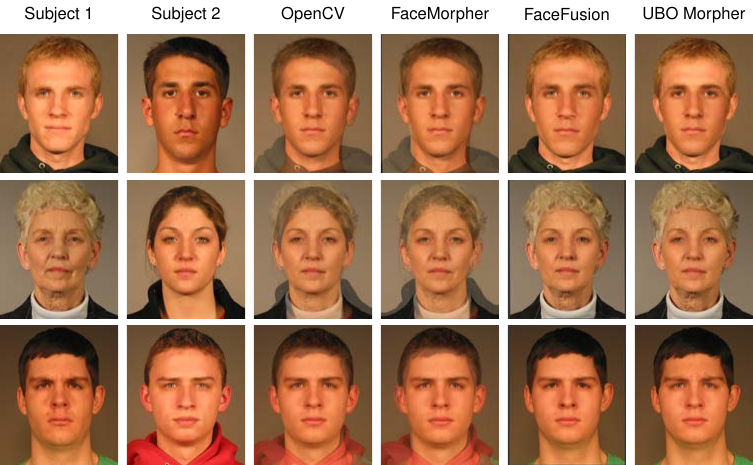}
    \caption{Examples of the different morphing tools and their corresponding bona fide subjects in the used dataset.}
    \label{fig:morphing_tools_examples}
\end{figure*}

Examples of morphed images and their contributing subjects are shown in~\autoref{fig:morphing_tools_examples} and a full overview of the used datasets is provided in~\autoref{tab:generated_database}.

\begin{table}[H]%
\caption{Overview of the number of bona fide and morph images in the used databases. The reference and probe images are bona fide and the morphed images are created from the reference images.}
\begin{adjustbox}{max width=\columnwidth}
\begin{tabular}{@{\extracolsep{2pt}}lcccccc@{}} \toprule 
     \textbf{Database} &  \textbf{Reference} & \textbf{Probe} & \textbf{OpenCV} & \textbf{FaceMorpher} & \textbf{FaceFusion}  & \textbf{UBO Morpher}   \\ \midrule
    FERET  & 622 & 786 & 529 & 529 & 529 & 529 \\
    FRGCv2 & 1441 & 1725 & 964  & 964 & 964 & 964 \\
    \bottomrule
    \end{tabular}
    \end{adjustbox}
\label{tab:generated_database}
\end{table}

The reason why only the datasets in~\cite{Scherhag-PRNU-TBIOM-2019} are considered and not other datasets, \eg from~\autoref{tab:fr_vulnerability_to_morphs}, is that they are either private or do not contain sufficient intra-class variations for realistically evaluating both the morphing vulnerability and the face recognition accuracy. Many morphing databases are derived from the Face Research London Lab (FRLL) dataset~\cite{debruine2017face}, \eg~\cite{Damer-MorDIFF-IWBF-2023, Sarkar-AreGANsThreateningFR-ICASSP-2022} and in the FRLL dataset, bona fide reference and probe images are considered from within the same capturing session with only minor differences in expression. This is also the case for many images in FERET, but FERET contains significantly more subjects and also has some intra-class variations (see \eg \autoref{fig:feret_ref_img} and \ref{fig:feret_probe_img}).

\subsection{Training Details}
The models were trained for $100$ epochs using a batch size of 256, stochastic gradient descent with a learning rate of $1e-1$, and Nesterov momentum~\cite{Surskever-ImportanceOfInitializationAndMomentumInDL-ConfMachineLearning-2013} of $0.9$. Additionally, a weight decay of $1e-1$ was added each $3$ epochs with a early stopping patience at $10$ epochs. 

\subsection{Experimental Protocols}
To evaluate the proposed method, experiments are carried out by training on the FERET database using the OpenCV and FaceMorpher morphing tools as training and UBO Morpher as validation. For evaluation, we used the UBO Morpher and FaceFusion morphing subsets of the FRGCv2 dataset. This represents the most realistic scenario as FRGCv2 contains realistic intra-class variations (see~\autoref{fig:reference_probe_samples}), and since we evaluate using the two most realistic morphing tools (see~\autoref{fig:morphing_tools_examples}). The same training and evaluation protocols are repeated separately for each tested face recognition system and backbone. 

\subsection{Metrics}

The face recognition performance will be evaluated according to the False Match Rate (FMR) and False Non-Match Rate (FNMR) as further described in~\cite{ISO-IEC-19795-1-Framework-210216}. Furthermore, the vulnerability of a system is measured in terms of the Impostor Attack Presentation Accept Rate (IAPAR) which for a full-system evaluation is defined as the \textit{proportion of impostor attack presentations using the same presentation attack instrument (PAI) species that result in an accept}~\cite{ISO-IEC-30107-3-PAD-metrics-2023}. As discussed in ISO/IEC 30107-3:2023~\cite{ISO-IEC-30107-3-PAD-metrics-2023} the problem of only looking at the IAPAR at high security values, \eg FMR=0.1\% is that the convenience of the system in terms of FNMR is not considered. Hence, following the suggestion in~\cite{Scherhag-MorphingAttacks-MorphingTechniques-BIOSIG-2017}, ISO/IEC 30107-3:2023 defined that the RIAPAR metrics should be used to evaluate the generalized performance of a face recognition system to morphing attacks. Given a threshold, $\tau$, the RIAPAR is defined as:

\begin{equation}
RIAPAR(\tau) = FRR(\tau) + IAPAR(\tau)
\end{equation}

where in our case the false reject rate (FRR) is equal to the FNMR of the system. 

Specifically, we evaluate the RIAPAR by considering high security thresholds as recommended by FRONTEX~\cite{FRONTEX2015}. FRONTEX recommends that for border security applications, a threshold corresponding to FMR=0.1\% is used. Hence, we evaluate RIAPAR at thresholds corresponding to FMR=0.1\%, FMR=0.01\%, and FMR=0.001\%. During the experiments, the thresholds corresponding to the above FMR values are calculated from scores obtained by doing more than two million bona fide non-mated comparisons on the FRGCv2 database.

\section{Generalized Full-System Evaluation}
\label{sec:generalized_system_evaluation}

\begin{table}[t!]
\caption{Measures for a generalized full system evaluation when trained on FERET and tested on FRGC using the Original and TetraLoss embeddings. The used MAD subsystem is from~\cite{Scherhag-FaceMorphingAttacks-TIFS-2020}. Best RIAPAR for each operating point and face recognition system is marked in \textbf{bold}.}
\label{tab:generalized_full_system_evaluation_frgc}
\begin{subtable}{\columnwidth}
\begin{adjustbox}{max width=\columnwidth}
\begin{tabular}{@{\extracolsep{2pt}}llccc@{}} \toprule 
\multicolumn{1}{c}{}  &  \multicolumn{1}{c}{}  & \multicolumn{3}{c}{\textbf{RIAPAR}} \\ \cmidrule{3-5}
\multirow{-2}{*}{\textbf{System}} & \multirow{-2}{*}{\textbf{Scenario}}  & \textbf{FMR=0.1$\%$}      & \textbf{FMR=0.01$\%$}  & \textbf{FMR=0.001$\%$}  \\ \midrule
\multirow{4}{*}{ArcFace} 
& Original & 0.76 & 0.58 & 0.28    \\
& Original \& MAD & 0.76 & 0.58 & 0.28     \\
& Tetra  & \textbf{0.28} & 0.17 & 0.16       \\
& Tetra \& MAD & \textbf{0.28} & \textbf{0.16} & \textbf{0.10} \\ \midrule
\multirow{4}{*}{MagFace} 
& Original &  0.75 &  0.56 &  0.25       \\
& Original \& MAD & 0.75 & 0.56 & 0.25    \\
& Tetra  & \textbf{0.28} & 0.18 & 0.17 \\ 
& Tetra \& MAD & \textbf{0.28} & \textbf{0.16} & \textbf{0.10} \\ \midrule
\multirow{4}{*}{AdaFace} 
& Original  & 0.77 & 0.60 & 0.32     \\
& Original \& MAD & 0.77 & 0.60 & 0.31     \\
& Tetra  & \textbf{0.28} &  0.18 & 0.18        \\
& Tetra \& MAD & \textbf{0.28} & \textbf{0.17} & \textbf{0.10} \\  \bottomrule
\end{tabular}
\end{adjustbox}
\vspace{0.05cm}
\caption{ResNet100}
\end{subtable}

   \begin{subtable}{\columnwidth}
\label{tab:iapmr_tshirt_attacks}
\begin{adjustbox}{max width=\columnwidth}
\begin{tabular}{@{\extracolsep{2pt}}llccc@{}} \toprule 
\multicolumn{1}{c}{}  &  \multicolumn{1}{c}{}  & \multicolumn{3}{c}{\textbf{RIAPAR}} \\ \cmidrule{3-5}
\multirow{-2}{*}{\textbf{System}} & \multirow{-2}{*}{\textbf{Scenario}}  & \textbf{FMR=0.1$\%$}      & \textbf{FMR=0.01$\%$}  & \textbf{FMR=0.001$\%$}  \\ \midrule
\multirow{4}{*}{ArcFace} 
& Original &  0.76 & 0.57 & 0.25   \\
& Original \& MAD & 0.76  & 0.57 & 0.26    \\
& Tetra  & 0.26 & 0.21 & 0.32         \\
& Tetra \& MAD & \textbf{0.25} & \textbf{0.18} & \textbf{0.19} \\ \midrule
\multirow{4}{*}{MagFace} 
& Original &  0.71 & 0.53 & 0.29       \\
& Original \& MAD & 0.71 & 0.53 & 0.29    \\
& Tetra  & 0.26 & 0.17 & 0.23        \\ 
& Tetra \& MAD & \textbf{0.25} & \textbf{0.14} & \textbf{0.10} \\ \midrule
\multirow{4}{*}{AdaFace} 
& Original  & 0.76 & 0.60 & 0.25    \\
& Original \& MAD & 0.76 & 0.59 & 0.25    \\
& Tetra  & 0.26 & 0.20 & 0.22        \\
& Tetra \& MAD & \textbf{0.25} & \textbf{0.16} & \textbf{0.12} \\\bottomrule
\end{tabular}
\end{adjustbox}
\vspace{0.05cm}
\caption{ResNet50} \vspace{-0.4cm}
\end{subtable}
\end{table}

Table~\ref{tab:generalized_full_system_evaluation_frgc} shows the RIAPAR scores at different operational relevant operating points when training on FERET and evaluating on FRGCv2 across four different scenarios. In the \emph{Original} scenario the embeddings from the pre-trained models provided by the authors of the corresponding works are used and for \emph{Tetra} the embeddings fine-tuned from the original systems using TetraLoss is used. For the other two cases, we combine the two face recognition systems with the MAD algorithm from~\cite{Scherhag-FaceMorphingAttacks-TIFS-2020}. This algorithm works by training a SVM over the difference vectors obtained when subtracting a probe image from a suspected image. According to the benchmark in~\cite{NGAN-PerformanceOfAutomatedFaceMorphDetection-FRVT-2020}, it represents one of the best performing D-MAD algorithms. We train the algorithm on the embeddings obtained from the pre-trained face recognition systems and backbones. While we do not show the performance of the baseline detection algorithms we trained, we ensured that the detection equal error rates (D-EERs) are similar to the D-EERs reported in the original work~\cite{Scherhag-FaceMorphingAttacks-TIFS-2020}. One way to combine a detection system and face recognition system is using score fusion as suggested in ~\cite{Chingovska-BiometricEvaluation-TIFS-2014, Ibsen-DigitalFaceManipulationsBiometricSystems-Handbook-2022}. Accordingly, in this work we perform a simple average fusion over the scores. While this might not be the optimal fusion approach, it still provides a valid baseline for comparing the proposed TetraLoss against the original system when used in conjunction with a MAD system. Moreover, a more advanced weighting would have required an additional morphing database for training. 

As seen in~\autoref{tab:generalized_full_system_evaluation_frgc} the RIAPAR scores obtained from the proposed method significantly outperform the original embeddings when evaluating the joint performance of the system in terms of vulnerability to morph and false reject rate. When comparing the original and proposed system without MAD, we can see that the RIAPAR of TetraLoss is better across almost every face recognition system and operating point. For instance, for FMR=0.1\% the RIAPAR improves with more than 45 percentage points for both ResNet50 and ResNet100 using ArcFace, MagFace, and AdaFace. Similarly, significant gains are observed for FMR=0.01\%. For FMR=0.001\%, the performance gains in terms of RIAPAR using ResNet100 is between $8-14$ percentage points whereas for ResNet50 it is between $3-6$ percentage points expect for ArcFace where the original system is better. When evaluating both systems together with MAD, the proposed method still significantly outperforms the original system. Furthermore, the proposed system even, in some cases, improves when combined with MAD, especially for FMR=0.001\%. The reason for this behaviour is that the used MAD system performs well on zero-effort (non-mated) attacks and hence complements the proposed system. Contrarily, the original system does not improve when adding the MAD module. The reason for this is that although the comparison scores for the morphs get lower for the fused system (less similar to the contributing subjects), the thresholds corresponding to the different FMRs also decrease and hence the system still accepts the morphed images. Since both the original face recognition system and MAD system work over the same embeddings this behavior makes sense. Future work could benchmark further baseline MAD algorithms; however, it should be noted that the MAD algorithm included in this work outperforms many existing algorithms, \eg methods which use texture descriptors such as local binary patterns as well as landmark-based approaches and demorphing~\cite{Scherhag-FaceMorphingAttacks-TIFS-2020}. 
It should also be noted that for all systems and backbones at the three selected operating points, the original system retains a FNMR close to 0\% whereas the FNMR of the TetraLoss system gradually increases and hence degrades performance on bona fide data (see \autoref{tab:feasibility_results}). In general, the FNMR of the original system will also increase if the security threshold is chosen based on even lower FMR values than evaluated herein (as shown in~\cite{Ibsen-DigitalFaceManipulationsBiometricSystems-Handbook-2022} on similar data). Future work could focus on more challenging databases with more intra-class variation and using a more significant number of morphed images. 
Nevertheless, the experiments in this work indicate that, at relevant operating points, the proposed method can improve the robustness of existing face recognition systems towards morphs and be used with an existing state-of-the-art MAD algorithm.

\begin{figure*}[!tb]
\begin{subfigure}[t]{0.24\linewidth}
  \includegraphics[width=\linewidth]{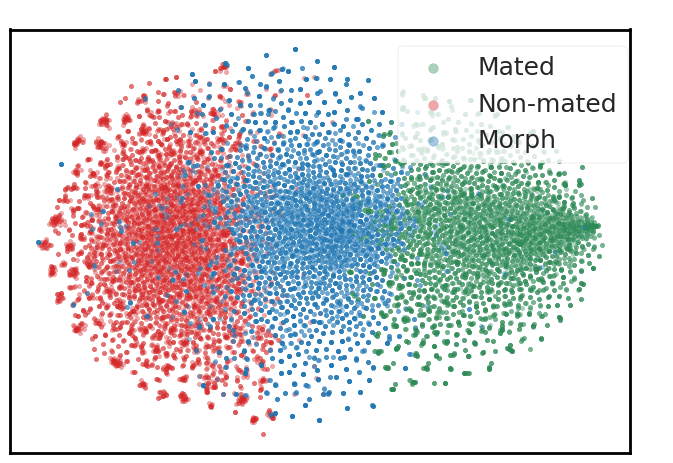}
  \caption{AdaFace}
\end{subfigure} %
\begin{subfigure}[t]{0.24\linewidth}
  \includegraphics[width=\linewidth]{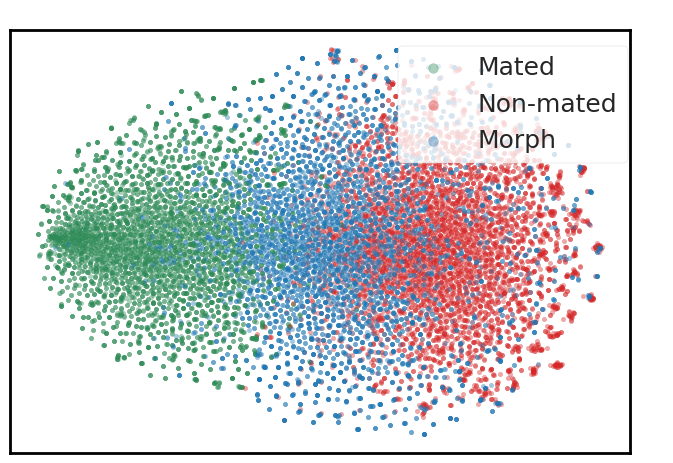}
  \caption{TetraLoss}
  \label{fig:tetra_resnet100}
\end{subfigure} %
\centering
\begin{subfigure}[t]{0.24\linewidth}
  \includegraphics[width=\linewidth]{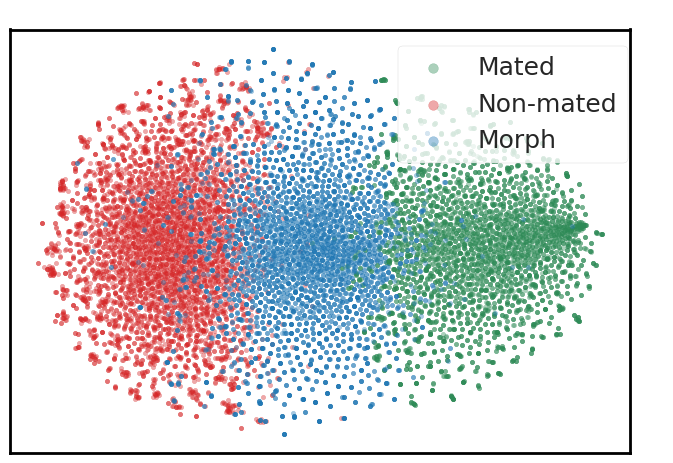}
  \caption{AdaFace}
\end{subfigure} %
\begin{subfigure}[t]{0.24\linewidth}
  \includegraphics[width=\linewidth]{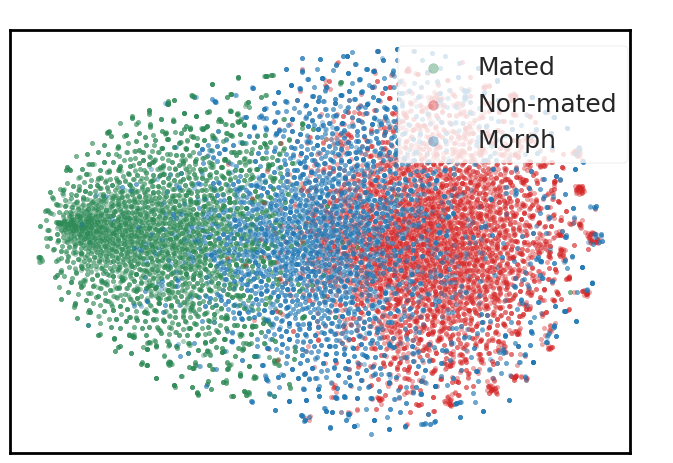}
  \caption{TetraLoss}
  \label{fig:tetra_resnet50}
\end{subfigure} %
\caption{t-SNE visualisation of the original AdaFace embeddings (a, c) and the modified TetraLoss embeddings (b, d) using ResNet100 (a, b) and ResNet50 (c, d).}
\label{fig:tsne-figures}
\end{figure*}

The improvements in terms of RIAPAR compared to the original system are achieved since TetraLoss is capable of pushing the embeddings of the morphed embeddings closer to the negative samples. To visualize this, consider two arbitrary embeddings, $x_a$ and $x_b$, used in a biometric comparison. We can then calculate the squared difference vector, $(x_a-x_b)^2$, between these two embeddings. \autoref{fig:tsne-figures} shows a t-SNE visualization over these difference vectors when computed on 1000 randomly selected mated, non-mated and morph comparisons of the original AdaFace and modified TetraLoss embeddings using ResNet100 and ResNet50 backbones. The shown difference vectors are computed using the exact comparisons. Note that TetraLoss, in contrast to the baseline systems, seems to push the morph difference embeddings (blue points) closer to the non-mated distributions (red points), which is the desired behaviour. Ideally, the non-mated and morph embeddings should be completely overlapping. The issue with the embeddings of the original system is that, although visually there is a good separation between the three types of comparisons, we must consider the operating configuration of the system. In such a case, the operating threshold of the system would be defined according to the FMR of the system, and hence, both morphed and mated comparisons would be accepted into the system (as shown by the high RIAPAR in~\autoref{tab:generalized_full_system_evaluation_frgc}). Obviously, more conservative decision thresholds could reduce the vulnerability towards morphing attacks. However, this would come at the cost of unacceptable high FNMR values~\cite{Ibsen-DigitalFaceManipulationsBiometricSystems-Handbook-2022}. Note also that the t-SNE spaces defined by TetraLoss (\ie \autoref{fig:tetra_resnet100} and \autoref{fig:tetra_resnet50}) still contain morph embeddings closer to the mated embeddings. Since TetraLoss is based on the triplet loss, it is limited by the number of quadruples that must be created to achieve optimal performance. In our experiments, the FERET training database consists of few subjects, \ie 529 subjects. Therefore, it is likely that the use of a larger training database can lead to a higher separation of the embedding spaces of mated samples and morph images.         

\section{Feasibility Study}
\label{sec:feasibility_results}
We showed in~\autoref{sec:generalized_system_evaluation} that the proposed method can significantly improve the robustness of state-of-the-art face recognition systems against morphing attacks. Here, we investigate the feasibility of the proposed TetraLoss function by comparing it to triplet loss (see~\autoref{eq:triplet_loss}). Since the existing face recognition systems are already trained to achieve high  intra-class and low inter-class similarity there is no point in using triplet loss in its original form as it will not increase the robustness of the system to morphing attacks. Hence, in this study we evaluate the case where triplet loss is used to push away the morphing embedding from a reference and probe image of one of the contributing subjects. Following the notation in~\autoref{sec:proposed_framework}, $x^n_i$ is replaced by $x^m_i$ in~\autoref{eq:triplet_loss}, where $x^m_i$ represent the embedding of a morphed image such that one of the contributing subjects is the same as the subject in the anchor and positive embeddings. 

\begin{table*}[!tb]
\centering
\caption{Measures for a generalized full-system evaluation when trained on FERET and tested on FRGC using the triplet loss and TetraLoss embeddings.  Best RIAPAR for each operating point and face recognition system is marked in \textbf{bold}.}
\label{tab:feasibility_results}
\begin{subtable}{1\textwidth}
\centering
\begin{adjustbox}{max width=\linewidth}
\begin{tabular}{@{\extracolsep{2pt}}llccccccccc@{}} \toprule 
\multicolumn{1}{c}{}  &  \multicolumn{1}{c}{}  & \multicolumn{3}{c}{\textbf{FMR=0.1$\%$}} & \multicolumn{3}{c}{\textbf{FMR=0.01$\%$}} & \multicolumn{3}{c}{\textbf{FMR=0.001$\%$}}\\ \cmidrule{3-5} \cmidrule{6-8} \cmidrule{9-11} 
\multirow{-2}{*}{\textbf{System}} & \multirow{-2}{*}{\textbf{Scenario}}  & \textbf{IAPAR}      & \textbf{FNMR}  & \textbf{RIAPAR}  & \textbf{IAPAR}      & \textbf{FNMR}  & \textbf{RIAPAR}   & \textbf{IAPAR}      & \textbf{FNMR}  & \textbf{RIAPAR}  \\ \midrule
\multirow{4}{*}{ArcFace} 
& Triplet & 0.08 & 0.17 & 0.25 & 0.03 & 0.36 & 0.39 & 0.01 & 0.56 & 0.57     \\
& Triplet$_2$ &  0.08 & 0.17 & \textbf{0.24} &  0.02 & 0.42 & 0.44 &  0.01 & 0.67 & 0.68    \\
& Tetra &  0.27 & 0.01 & 0.28 &  0.14 & 0.03 & \textbf{0.17} &  0.06 & 0.10 & \textbf{0.16}          \\
& Tetra$_2$ & 0.28 & 0.00 & 0.29 &  0.14 & 0.02 & \textbf{0.17} &  0.05 & 0.11 & \textbf{0.16}        \\  \midrule
\multirow{4}{*}{MagFace} 
& Triplet &   0.09 & 0.16 & \textbf{0.25} &  0.03 & 0.37 & 0.39 &  0.01 & 0.60 & 0.60     \\
& Triplet$_2$ &  0.05 & 0.28 & 0.33 &  0.01 & 0.66 & 0.67 &  0.00 & 0.86 & 0.87     \\
& Tetra & 0.27 & 0.01 & 0.28 &  0.14 & 0.04 & 0.18 &  0.06 & 0.11 & \textbf{0.17}         \\
& Tetra$_2$ & 0.24 & 0.01 & \textbf{0.25} &  0.09 & 0.07 & \textbf{0.15} &  0.02 & 0.25 & 0.27         \\ \midrule
\multirow{4}{*}{AdaFace} 
& Triplet &  0.14 & 0.08 & \textbf{0.22} &  0.05 & 0.21 & 0.26 &  0.01 & 0.46 & 0.47   \\
& Triplet$_2$ & 0.17 & 0.05 & \textbf{0.22} &  0.08 & 0.13 & 0.21 &  0.03 & 0.27 & 0.30     \\
& Tetra &  0.27 & 0.01 & 0.28 &  0.15 & 0.04 & \textbf{0.18} &  0.05 & 0.12 & 0.18         \\
& Tetra$_2$ & 0.28 & 0.01 & 0.29 & 0.16 & 0.03 & 0.19 &  0.06 & 0.11 & \textbf{0.17}        \\ \midrule
\end{tabular}
\end{adjustbox}
\caption{ResNet100}
\end{subtable}

\begin{subtable}{\textwidth}
\centering
\begin{adjustbox}{max width=\linewidth}
\begin{tabular}{@{\extracolsep{2pt}}llccccccccc@{}} \toprule 
\multicolumn{1}{c}{}  &  \multicolumn{1}{c}{}  & \multicolumn{3}{c}{\textbf{FMR=0.1$\%$}} & \multicolumn{3}{c}{\textbf{FMR=0.01$\%$}} & \multicolumn{3}{c}{\textbf{FMR=0.001$\%$}}\\ \cmidrule{3-5} \cmidrule{6-8} \cmidrule{9-11} 
\multirow{-2}{*}{\textbf{System}} & \multirow{-2}{*}{\textbf{Scenario}}  & \textbf{IAPAR}      & \textbf{FNMR}  & \textbf{RIAPAR}  & \textbf{IAPAR}      & \textbf{FNMR}  & \textbf{RIAPAR}   & \textbf{IAPAR}      & \textbf{FNMR}  & \textbf{RIAPAR}  \\ \midrule
\multirow{4}{*}{ArcFace} 
& Triplet & 0.08 & 0.23 & 0.31 & 0.02 & 0.44 & 0.46 & 0.01 & 0.63 & 0.64   \\
& Triplet$_2$ &0.14 & 0.10 & \textbf{0.24} & 0.06 & 0.25 & 0.31 & 0.02 & 0.43 & 0.46    \\
& Tetra & 0.22 & 0.04 & 0.26 & 0.11 & 0.11 & \textbf{0.21} & 0.03 & 0.29 & 0.32    \\
& Tetra$_2$ & 0.25 & 0.03 & 0.27 & 0.13 & 0.08 & \textbf{0.21} & 0.06 & 0.18 & \textbf{0.24}    \\  \midrule
\multirow{4}{*}{MagFace} 
& Triplet & 0.11 & 0.15 & 0.26 & 0.04 & 0.36 & 0.40 & 0.01 & 0.55 & 0.56     \\
& Triplet$_2$ & 0.12 & 0.11 & \textbf{0.24} & 0.04 & 0.29 & 0.33 & 0.01 & 0.53 & 0.54     \\
& Tetra & 0.25 & 0.01 & 0.26 &  0.11 & 0.06 & \textbf{0.17} & 0.04 & 0.19 & 0.23 \\
& Tetra$_2$ & 0.26 & 0.01 & 0.28 & 0.14 & 0.05 & 0.19 & 0.05 & 0.15 & \textbf{0.20}        \\ \midrule
\multirow{4}{*}{AdaFace} 
& Triplet & 0.11 & 0.15 & 0.26 & 0.05 & 0.30 & 0.34 & 0.02 & 0.47 & 0.48   \\
& Triplet$_2$ & 0.14 & 0.09 & \textbf{0.24} & 0.06 & 0.21 & 0.27 &  0.03 & 0.38 & 0.40     \\
& Tetra & 0.24 & 0.02 & 0.26 & 0.12 & 0.07 & 0.20 & 0.06 & 0.16 & 0.22        \\
& Tetra$_2$ & 0.26 & 0.01 & 0.27 & 0.14 & 0.05 & \textbf{0.19} & 0.06 & 0.14 & \textbf{0.20}        \\ \midrule
\end{tabular}
\end{adjustbox}
\caption{ResNet50} %
\end{subtable}
\end{table*}

To analyse the above in more detail, we expand the definition of triplet loss and TetraLoss as defined in~\autoref{eq:triplet_loss} and~\ref{eq:tetraloss}, respectively. Specifically, consider the set of subjects $S$ and any morphed sample $x^{m}_i \in X$. Then let $s_1, s_2 \in S$ be the subjects contributing to the morph $x^{m}_i$. In this case, we can find anchors and positives  $x^{a1}_i, x^{a2}_i,  x^{p1}_i,  x^{p2}_i \in X$ such that $x^{a1}_i$ and $x^{p1}_i$ are samples belonging to subject $s_1$ whereas $x^{a2}_i$ and $x^{p2}_i$ belongs to $s_2$. Furthermore, we can find a negative sample $x^n_i$ which neither belongs to $s_1$ nor to $s_2$. Using this logic, we can then expand the definition of triplet and TetraLoss in~\autoref{eq:triplet_loss} and~\ref{eq:tetraloss} such that $L_{Triplet_1}$ and $L_{Tetra_1}$ are defined over the triplets ($x^{a1}_i, x^{p1}_i, x^{m}_i$) and quadruplet ($x^{a1}_i, x^{p1}_i, x^{m}_i, x^{n}_i$), respectively. Similar $L_{Triplet_2}$ and  $L_{Tetra_2}$ are defined over the triplets ($x^{a2}_i, x^{p2}_i, x^{m}_i$) and quadruplet ($x^{a2}_i, x^{p2}_i, x^{m}_i, x^{n}_i$), respectively. 

In this case we define the following experimental scenarios:

    \begin{enumerate}
        \item Triplet: $L_{Triple_1}$ is calculated for each batch \vspace{0.1cm}
        \item Tetra: $L_{Tetra_1}$ is calculated for each batch
        \vspace{0.1cm}
        \item Triplet$_2$: $\frac{L_{\text{Triple}_1} + L_{\text{Triple}_2}}{2}$ is calculated for each batch
        \vspace{0.1cm}
        \item Tetra$_2$:  $\frac{L_{\text{Tetra}_1} + L_{\text{Tetra}_2}}{2}$ is calculated for each batch
        \vspace{0.1cm}
    \end{enumerate}

The results when training on FERET and testing on FRGCv2 for ResNet100 and ResNet50 are shown in~\autoref{tab:feasibility_results}. Note that the use of TetraLoss over the triplet loss achieves, in general, the best performance in terms of FNMR and RIAPAR, especially for high security thresholds. As expected, the IAPAR using triplet loss is in general better. However, a problem with triplet loss is that the FNMR rate becomes very high, \eg 17\% at a FMR=0.1\% for ArcFace using ResNet100. This indicates the case where one out of a thousand is falsely accepted, while 17\% of the mated comparisons are rejected by the system. The FNMR increases to 0.56 and 0.67 at FMR=0.001\% for the Triplet and Triplet$_2$ scenarios, respectively. Such a high number of false non-matches would make the system inapplicable in crowded border control situations, as it would require manual checks of the documents. Contrarily, for ArcFace using ResNet100, both TetraLoss variants have an FNMR of approximately 1\% or lower at FMR=0.1\%. Similar trends is observable for MagFace and AdaFace using the ResNet100 and ResNet50 backbones.

\section{CONCLUSION}
\label{sec:conclusion}
In this paper, we proposed a novel framework and loss function to make existing deep learning-based face recognition systems more robust against morphing attacks. To this end, a neural network architecture was optimised by a novel TetraLoss function to learn to separate morphing attack embeddings from bona fide embeddings extracted from face images of the subjects contributing to the morphed samples. The results show that the proposed method can seamlessly be applied on top of existing state-of-the-art face recognition systems to increase their resilience to morphing attacks while still achieving good face recognition performance on bona fide data. Specifically, using two distinct backbone architectures, the RIAPAR improves with at least 45\% points for the operational relevant threshold at FMR=0.1\% for ArcFace, MagFace, and AdaFace.

\section{Acknowledgements}

This research work has been partially funded by the Hessian Ministry of the Interior and Sport in the course of the Bio4ensics project, by German Federal Ministry of Education and Research and the Hessian Ministry of Higher Education, Research, Science and the Arts within their joint support of the National Research Center for Applied Cybersecurity, ATHENE and the European Union's Horizon 2020 research and innovation programme under the Marie Sk\l{}odowska-Curie grant agreement No. 860813 - TReSPAsS-ETN and by the European Union’s Horizon 2020 research and innovation program under grant agreement No. 883356 - iMARS.

{\small
\bibliographystyle{ieee}
\bibliography{egbib}
}
\end{document}

%% file: packages.tex
\usepackage{float}
\usepackage{multirow, multicol}
\usepackage{graphicx}
\usepackage{hhline}
\usepackage{amsmath}
\usepackage{array, makecell}%
\usepackage{bm}
\usepackage{amsbsy}
\usepackage{xspace}
\usepackage{xcolor}
\usepackage{newfloat}
\usepackage[export]{adjustbox}
\usepackage{enumitem}
\usepackage{url}
\usepackage{subcaption}
\usepackage{float}
\usepackage{graphicx}
\usepackage{hhline}
\usepackage{amsmath}
\usepackage{array, makecell}%
\usepackage{bm}
\usepackage{amsbsy}
\usepackage{booktabs}
\usepackage{xspace}
\usepackage{xcolor}
\usepackage{newfloat}
\usepackage[export]{adjustbox}
\usepackage{enumitem}
\usepackage{url}
\usepackage{times}
\usepackage{epsfig}
\usepackage{graphicx}
\usepackage{amsmath}
\usepackage{amssymb}
\usepackage{tabularx}
\usepackage[hidelinks]{hyperref}
\usepackage{multirow, multicol}
\usepackage{tikz}

%% file: commands.tex
\def\eg{{e.g.,}\@\xspace} 
\def\ie{{i.e.,}\@\xspace}